\documentclass{article}
\usepackage{spconf,amsmath,graphicx}
\usepackage{booktabs}
\usepackage{url}

\title{LWIRPOSE: A novel LWIR Thermal Image Dataset and Benchmark}
\name{Avinash Upadhyay$^{\dagger}$\quad  Bhipanshu Dhupar$^{\ddagger}$ \quad Manoj Sharma$^{\S}$\quad Ankit Shukla$^{\ddagger}$ \quad Ajith Abraham$^{\P}$}
\address{ $^{\dagger}$ $^{\ddagger}$ $^{\S}$ $^{\ddagger}$ $^{\P}$ Bennett University, Greater Noida , $^{\ddagger}$ Visual Cognition Lab. Pvt. Ltd. 
\\  $^{\dagger}$ avinres@gmail.com, $^{\S}$ mksnith@gmail.com,$^{\ddagger}$ anktshkla.1307@gmail.com}
\begin{document}
%
\maketitle
\begin{abstract}
Human pose estimation faces hurdles in real-world applications due to factors like lighting changes, occlusions, and cluttered environments. We introduce a unique RGB-Thermal Nearly Paired and Annotated 2D Pose Dataset, comprising over 2,400 high-quality LWIR (thermal) images. Each image is meticulously annotated with 2D human poses, offering a valuable resource for researchers and practitioners. This dataset, captured from seven actors performing diverse everyday activities like sitting, eating, and walking, facilitates pose estimation on occlusion and other challenging scenarios. We benchmark state-of-the-art pose estimation methods on the dataset to showcase its potential, establishing a strong baseline for future research. Our results demonstrate the dataset's effectiveness in promoting advancements in pose estimation for various applications, including surveillance, healthcare, and sports analytics. The dataset and code are available at \url{https://github.com/avinres/LWIRPOSE}
\end{abstract}
\begin{keywords}
Human Pose Estimation, Thermal Imagery, Thermal 2D pose estimation
\end{keywords}
\section{Introduction}
\label{sec:intro}
Thermal imaging technology captures temperature differences in the environment, which allows for detecting humans even in low-light conditions or through smoke, fog, or other visual obstructions. This makes thermal imaging particularly useful for military and defense applications where situational awareness is critical. For instance, 2D human pose estimation on thermal images can be used to track the movements of soldiers and other objects, enabling more accurate targeting and improved battlefield management. Additionally, thermal imaging can be employed in search and rescue missions, allowing rescuers to locate survivors in burning buildings or under debris. Besides military and defence applications, 2D human pose estimation on thermal images has numerous civilian uses. For example, it can be utilized in healthcare to monitor patients' vital signs and detect abnormalities in their body posture in darkness \cite{liu2019seeing}. In sports analytics, it can be applied to assess athletes' performance and optimize their training regimens. Furthermore, smart home automation can enable intelligent systems to recognize and respond to users' gestures and movements.

Despite the potential benefits of 2D human pose estimation on thermal images, there is a conspicuous absence of research in this area. Most existing works \cite{xu2022vitpose+, xu2022vitpose, openpose19, cheng2020higherhrnet} focus on visible light images, leaving thermal imaging. As a result, there is limited understanding of the challenges specific to 2D human pose estimation on thermal images and how to address them effectively.

The primary obstacle lies in the scarcity of large and diverse IR pose estimation datasets. Unlike the abundance of datasets like COCO \cite{lin2015microsoft} and MPII Human Pose \cite{mpii_dataset} for RGB images, IR datasets remain limited in size and scope. This hinders the training of robust and generalizable models, as they lack sufficient data to capture the full spectrum of human poses and variations. Additionally, IR images inherently differ in appearance and illumination compared to RGB. They lack the rich textures and color information crucial for traditional pose estimation algorithms, relying on thermal signatures that are highly variable depending on body temperature, clothing, and ambient environment. This variability in appearance and illumination makes it difficult for models to extract reliable features for accurate pose estimation. Compared to the distinct edges and textures defining body contours in RGB, IR images often exhibit blurry outlines and low contrast between body parts and the background. This ambiguity in visual features makes it difficult for models to accurately identify keypoints and differentiate between limbs in complex poses. Finally, human bodies naturally self-occlude during movement, posing a challenge even in RGB images. However, self-occlusion becomes even more problematic in IR due to the lack of clear visual cues. Thermal signatures can overlap, making it difficult to distinguish between limbs in contact, posing a significant challenge for models to accurately infer the underlying pose structure.

\begin{figure*}
    \centering
    \includegraphics[width=1\linewidth]{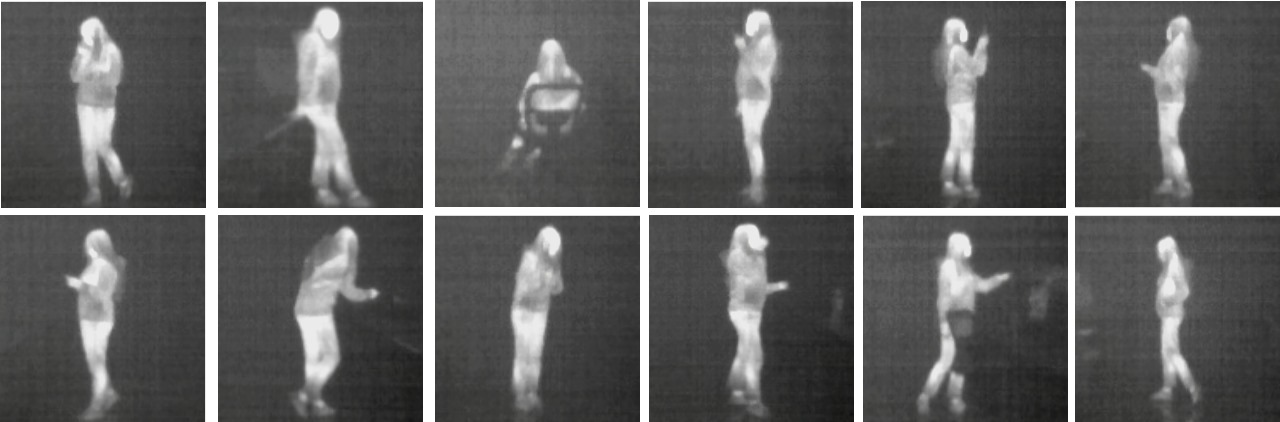}
    \caption{Sample of thermal images from the LWIRPose Dataset. The samples belongs to one subject performing 12 different activities. It is visible from the images that the data constitutes complexities such as occlusion, self-occlusion and noises.}
    \label{fig:sample1}
\end{figure*}
\begin{figure*}
    \centering
    \includegraphics[width=1\linewidth]{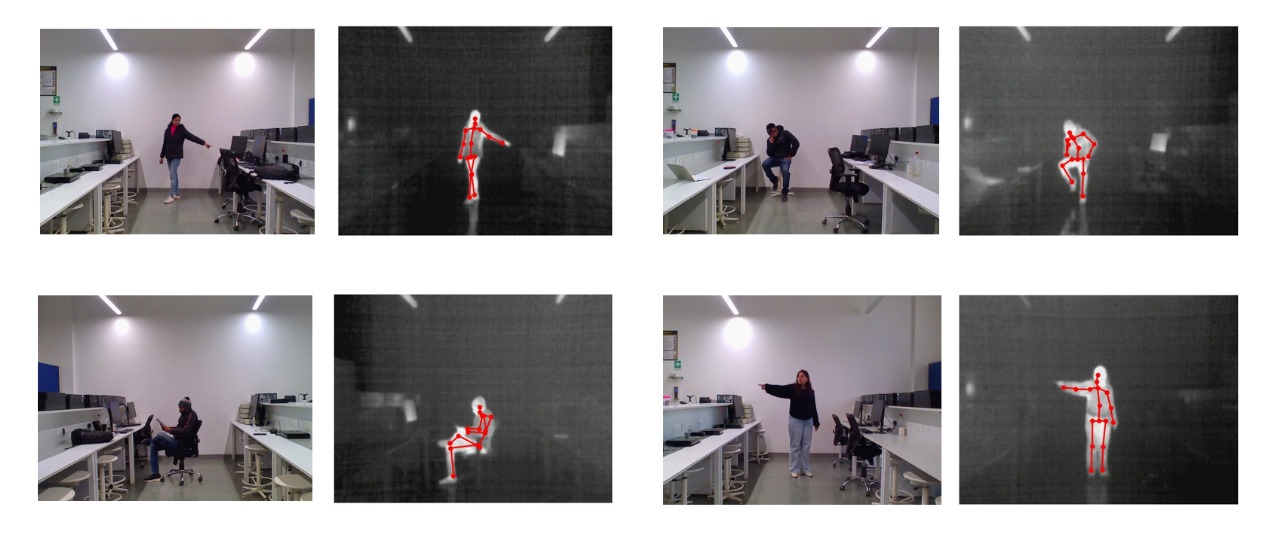}
    \caption{Image represents samples from the dataset. The RGB-IR images that were nearly paired were captured using the camera. The annotated pose points are shown on the LWIR images. Four different Subjects are performing different activities.}
    \label{fig:dataset_sample}
\end{figure*}

To address these challenges, the presented dataset introduces deliberate complexities that reflect real-world scenarios as shown in figure \ref{fig:sample1} \& \ref{fig:dataset_sample}. It encompasses a diverse range of: 
\begin{itemize}
    \item Pose variations: The dataset includes a variety of poses, from simple standing positions to complex actions like walking and greeting, ensuring the model can handle diverse scenarios.
    \item Body shapes: By including subjects with different body types, the dataset encourages the model to learn pose estimation independent of body shape variations.
    \item Clothing: The presence of diverse clothing styles helps the model learn pose estimation without relying solely on skin texture, which can be obscured by clothing in IR images.
    \item Self-occlusion: The inclusion of poses with self-occlusion challenges the model to infer pose even when parts of the body are hidden from view.
    \item Different activities: Incorporating various activities like sitting, walking, and exercise exposes the model to a broader range of motion patterns, improving its generalizability.
\end{itemize}
By incorporating these complexities, the dataset aims to train models that are robust to the challenges inherent in IR pose estimation and can perform accurately in real-world applications. While the effectiveness of the dataset depends on its size, quality, and diversity, it represents a significant step towards overcoming the unique challenges of IR pose estimation and paving the way for more accurate and robust models in this domain.


Major contribution of the paper is:
\begin{itemize}
    \item A Unique and Diverse Dataset: We introduce a novel RGB-Thermal Nearly Paired and Annotated 2D Pose Dataset, boasting over 2,400 high-quality LWIR (thermal) images alongside corresponding near-paired RGB images. This extensive dataset captures seven individuals performing various daily activities like walking, sitting, eating, and more, encompassing a wide range of scenarios and motions. Each image is meticulously annotated with accurate 2D human poses in the MPII format, providing valuable ground truth for researchers and practitioners.

    \item Benchmarking RGB-Based Methods: To demonstrate the dataset's utility, we conduct a comprehensive evaluation of state-of-the-art RGB-based pose estimation methods. These methods, originally designed for RGB data, are fine-tuned on our novel dataset, establishing a robust baseline for future research endeavours. This evaluation highlights the effectiveness of our dataset in pushing the boundaries of pose estimation performance.

\end{itemize}

\section{Related Works}
\label{sec:related}

In this section, related works on datasets and thermal pose estimation are discussed. State-of-the-art pose estimation methods on RGB images and thermal images are also discussed.

\subsection{Thermal Pose Estimation Dataset}

Human pose estimation using RGB images has received extensive attention in recent years \cite{xu2022vitpose+, xu2022vitpose, openpose19, cheng2020higherhrnet}. Several datasets \cite{lin2015microsoft, mpii_dataset} in the RGB domain are now available in the public domain, which has been widely used by these algorithms to develop 2D human pose estimation (2DHPE) algorithms. However, research involving thermal data remains scarce due to the limited availability of publicly accessible datasets. This work addresses this gap by introducing a novel dataset specifically designed for thermal pose estimation.

While a few thermal pose datasets exist, they have limitations. Chen et al. \cite{chen2020} proposed a thermal and visible image pose estimation dataset in indoor environments. They collected nearly 24,000 images and annotated the thermal images using the pose points obtained by running state-of-the-art (SotA) models on RGB images. The same pose points are used as ground truth for the thermal image after scaling. However, the dataset has certain limitations. Firstly, the resolution of the thermal images is very low, and the pose annotation is not properly supervised. Smith et al. \cite{javiersmith2023} proposed the UCH-ThermalPose dataset, containing both indoor and outdoor thermal images with annotated key points. However, its size is relatively small, and it only targets static poses. The dataset contains nearly 600 images collected from different thermal image datasets and manually annotated by the authors.
Our dataset offers significant advancements over these datasets. It features:
\begin{itemize}
\item Larger size: Over 2,400 high-resolution RGB and IR images.
\item Diverse activities: Actors performing various daily actions (walking, sitting, eating, etc.).
\item Near-paired RGB counterparts: Facilitate comparison and analysis.
\end{itemize}

\subsection{2D Human Pose Estimation on RGB and Thermal Images}

Due to the availability of high-quality annotated 2D human pose estimation data in the RGB domain, several algorithms that provide state-of-the-art results were developed. Simple Baseline \cite{Xiao2018baseline} utilizes a ResNet architecture to extract features and uses deconvolutional layers to estimate poses. HR-Net \cite{sun2019deep} has proposed the generation of heatmaps using multi-resolution connection subnetworks in parallel and fusing features at multi-scale. HRNet and its variants have been widely used in 2D human pose estimation tasks. OpenPose \cite{openpose17, openpose19} is another state-of-the-art model that uses part affinity fields (PAF) for heatmap-based keypoint coordinate grouping. Different vision transformer-based \cite{vit} models were developed for pose estimation with the advent of transformer-based networks. ViT Pose \cite{xu2022vitpose, xu2022vitpose+} have utilized \cite{vit} to extract feature maps of a person present, then a CNN-based decoder was used to estimate the 2D pose points.

There has been a very limited number of works in thermal human pose estimation. \cite{javiersmith2023} proposed a dataset and, along with it, established a baseline using existing RGB 2D human pose estimation models. \cite{chen2020} proposed a CNN-based feature extraction and PAF-based decoding network for thermal pose estimation.

In this work, we also provide a baseline for our dataset by running different existing RGB-based 2D pose estimation networks.
\section{Methodology}
\label{sec:method}

\begin{table}[htbp]
  \centering
  \caption{LWIR Imager specification}
    \begin{tabular}{|l|l|}
    \toprule
    \textbf{Parameter} & \textbf{Specification} \\
    \midrule
    \textbf{Spectral Range} & 8µm -14µm \\
    \midrule
    \textbf{Resolution} & 640 x 480 for RGB \& IR \\
    \midrule
    \textbf{FoV} & 57°C \\
    \midrule
    \textbf{Temperature Range} & -40° to 330°C \\
    \midrule
    \textbf{Frame rate} & $<$ 9 Hz \\
    \midrule
    \textbf{Focus} & Fixed \\
    \midrule
    \textbf{Type} & Un-cooled \\
    \bottomrule
    \end{tabular}%
  \label{tab:addlabel}%
\end{table}%

\subsection{Dataset Preparation}
We leverage a Seek thermal camera with long-wave infrared sensor setup operating at 7-14mm to capture thermal signatures of human poses at a resolution of 640x480 pixels, ensuring detailed representation of each subjects. With the help of advanced features of the camera, we were able to capture both RGB and their corresponding IR images. Each image within the dataset is meticulously annotated with 17 keypoints, precisely pinpointing major body joints such as head, shoulders, elbows, wrists, hips, knees, and ankles. This comprehensive set of pose points enables accurate estimation of various postures and movements, laying the groundwork for robust model training and evaluation. Moreover, to enhance the dataset's diversity and real-world applicability, we captured images across 12 distinct activity classes like "Discussion," "Smoking," and "Walking." Random pose variations were captured within each class, resulting in a comprehensive representation of human movement.


While our dataset included both RGB and corresponding infrared (IR) images, directly applying the HRNet model \cite{sun2019deep} to IR images yielded unsatisfactory pose estimation results. To overcome this challenge, we adopted a two-step approach. First, we passed the RGB images through the HRNet model \cite{sun2019deep}, obtaining accurate keypoint predictions. Subsequently, we transferred these predicted keypoints to their corresponding IR images. These key points were inaccurate as there was some mismatch in the registration of the RGB-thermal image pair. Further, many RGB-IR image pairs were not naturally captured together, necessitating manual annotation of the entire dataset. We developed a custom annotation tool to streamline this process, hence achieving robust pose ground truth in the thermal domain.

\begin{table}[htbp]
  \centering
  \caption{The number of images and annotated 2D poses in the LWIRPose dataset. The table represents the total number of images used for training and testing. We have used six subjects (1 Male and 5 Females) with 2106 images for training and one subject (Male) with 355 images for testing.}
    \begin{tabular}{|cccc|}
    \toprule
    \multicolumn{1}{|c|}{\textbf{Pose}} & \multicolumn{1}{c|}{\textbf{Training}} & \multicolumn{1}{c|}{\textbf{Testing}} & \textbf{Total} \\
    \midrule
    \multicolumn{1}{|c|}{} & \multicolumn{1}{c|}{S2+S3+S4+S5+S6+S7} & \multicolumn{1}{c|}{S1} &  \\
    \midrule
    \textbf{Direction} & 155   & 25    & 180 \\
    \textbf{Discussion} & 160   & 25    & 185 \\
    \textbf{Eating} & 160   & 25    & 185 \\
    \textbf{Greeting} & 154   & 25    & 179 \\
    \textbf{Phone Talk} & 153   & 25    & 178 \\
    \textbf{Posing} & 132   & 25    & 157 \\
    \textbf{Purchases} & 151   & 25    & 176 \\
    \textbf{Sitting} & 280   & 55    & 335 \\
    \textbf{Smoking} & 154   & 25    & 179 \\
    \textbf{Taking photo} & 180   & 25    & 205 \\
    \textbf{Waiting} & 150   & 25    & 175 \\
    \textbf{Walking} & 277   & 50    & 327 \\
    \midrule
    \textbf{Total} & 2106  & 355   & 2461 \\
    \bottomrule
    \end{tabular}%
  \label{tab:data_info}%
\end{table}%

\subsection{Experimental Settings} 

We captured the whole data in a controlled indoor environment in which the camera remained fixed at a distance of 7 meters for every object, ensuring consistent thermal signature capture across the dataset. This approach of leveraging RGB information combined with custom annotation and controlled data acquisition enabled us to create a high-quality IR pose estimation dataset, paving the way for further research in this domain. Table \ref{tab:data_info} represents the number of images for each activity.

\subsection{Data Processing}

The collected and annotated data was processed to train the existing RGB-based 2D HPE models. We have segregated the training and testing sets based on the subjects. The Training images included images from subjects S2, S3, S4, S5, S6 and S7 performing various activities. Images of subject S1 are preserved for testing purposes. Since the data size is small, we have omitted the validation set. Random samples from the training set were used to validate the model. The results demonstrated in the paper were obtained on the testing set. The split between the training and testing data and the number of images for each set is presented in Table \ref{tab:data_info}.

\subsection{Evaluation Metric}

The models are trained and evaluated using \textbf{Mean Per Joint Position Error (MPJPE) loss in pixels}, widely used in the Pose estimation tasks. MPJPE is the mean Euclidean distance between the GT 2D Pose \(p_{k,i}\) and estimated 2D Pose \(\hat{p}_{k,i}\). To pre-train the encoder and decoder of the ResNet, the loss was calculated as:

\[
L_{res} = \tfrac{1}{n}\left [ \sum_{k=1}^{n} \left \| p_{k,i} - \hat{p}_{k,i} \right \|_{2} \right ]
\]

Where n is the number of joints in a pose. Lower MPJPE represents better model pose point predictions.

Further, the \textbf{Percentage of correct key points} (PCKh) metric was also used to determine the accuracy of the localization of different critical points with a given threshold. Specifically, PCKh@0.5 is used as a threshold, which is 50\% of the head bone link. A higher PCK value represents better model performance.

\section{Results and Discussion}
\label{sec:res}

\begin{figure*}[h]
    \centering
    \includegraphics[width=0.9\linewidth]{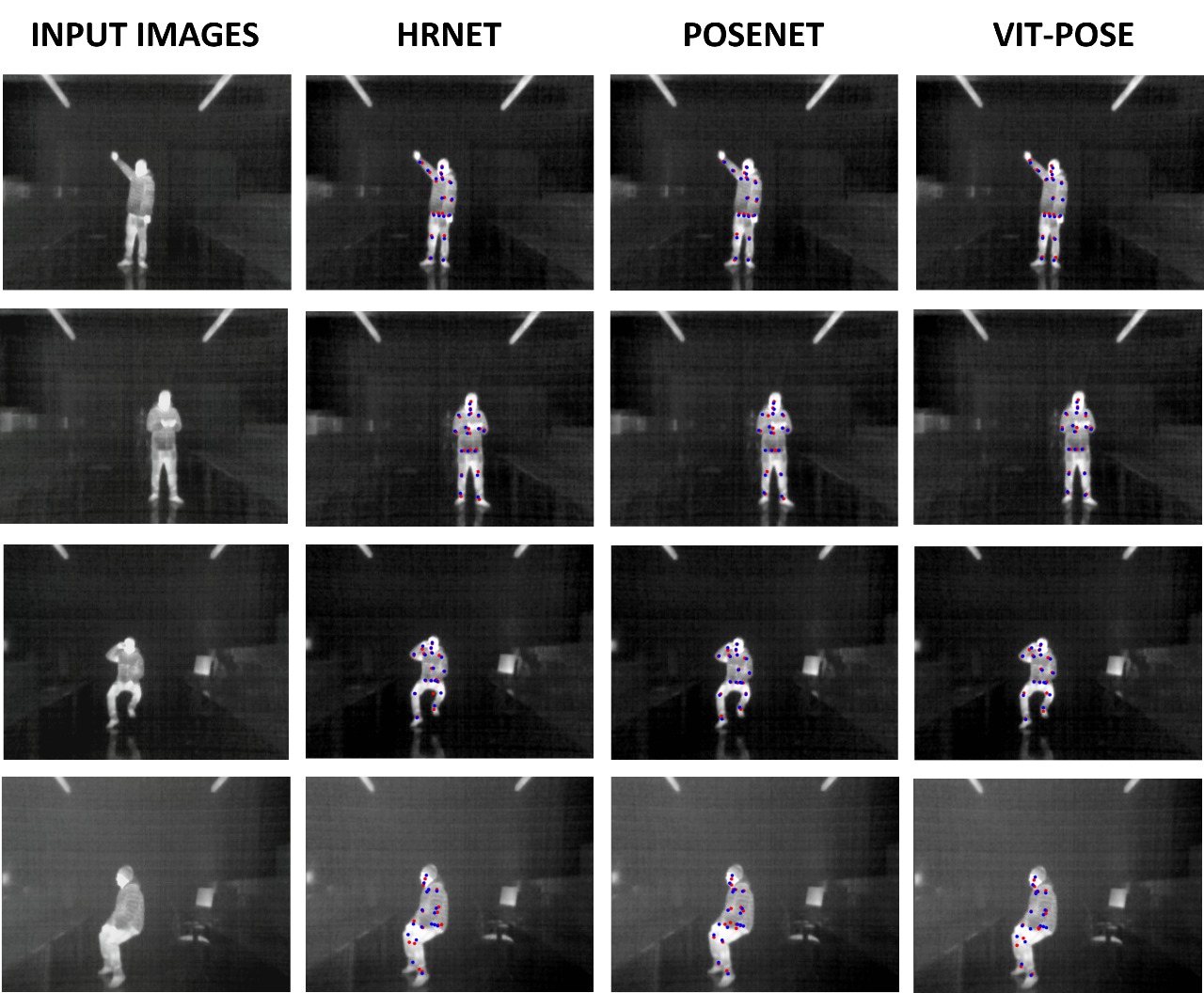}
    \caption{Visual Results of different Deep learning models on thermal images from the dataset. Blue Points are Ground Truth, and Red Points are Predicted Pose points. It can be seen that ViT-Pose has performed much better than other models. However, the performance of almost all existing RGB-based models deteriorates on the thermal images.}
    \label{fig:DL_results}
\end{figure*}

\begin{figure*}
    \centering
    \includegraphics[width=1\linewidth]{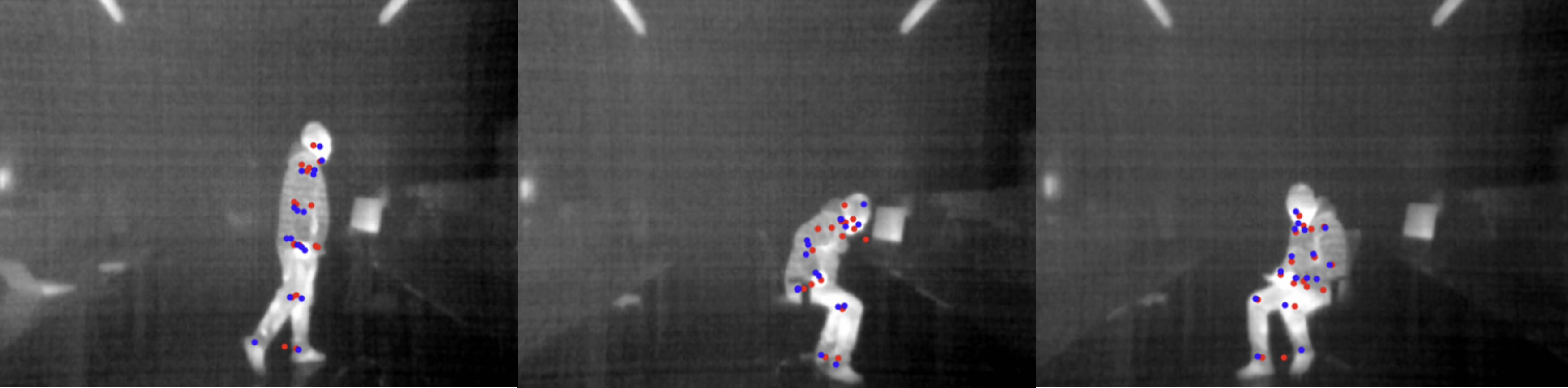}
    \caption{Failed cases of ViTPose. It can be seen that for complex poses ViTPose fails to extract and decode the features properly, representing the complexity involved with the LWIR images.}
    \label{fig:failed}
\end{figure*}

We have trained three different models ViTPose \cite{xu2022vitpose}, HRNet \cite{sun2019deep}, and simple baseline (ResNet-based pose estimation) \cite{Xiao2018baseline} to evaluate the performance of these models on our dataset. 


\begin{table*}[htbp]
  \centering
  \caption{This table presents the MPJPE errors for four different pose estimation models: ResNet50 baseline\cite{Xiao2018baseline}, ResNet152 baseline\cite{Xiao2018baseline}, HRNet\cite{sun2019deep}, and ViTPose\cite{xu2022vitpose}. The MPJPE values are provided for 12 different activity categories for each model, along with the average MPJPE calculated across all categories.}
  \resizebox{\textwidth}{!}{
    \begin{tabular}{|l|r|r|r|r|r|r|r|r|r|r|r|r|r|}
    \toprule
    \textbf{Methods} & \multicolumn{1}{l|}{\textbf{Direction}} & \multicolumn{1}{l|}{\textbf{Discussion}} & \multicolumn{1}{l|}{\textbf{Eating}} & \multicolumn{1}{l|}{\textbf{Greeting}} & \multicolumn{1}{l|}{\textbf{Phone}} & \multicolumn{1}{l|}{\textbf{Posing}} & \multicolumn{1}{l|}{\textbf{Purchases}} & \multicolumn{1}{l|}{\textbf{Sitting}} & \multicolumn{1}{l|}{\textbf{Smoking}} & \multicolumn{1}{l|}{\textbf{Photo}} & \multicolumn{1}{l|}{\textbf{Waiting}} & \multicolumn{1}{l|}{\textbf{Walking}} & \multicolumn{1}{l|}{\textbf{Total}} \\
    \midrule
    \textit{Resolution: 640x480} &       &       &       &       &       &       &       &       &       &       &       &       &  \\
    \midrule
    Baseline (ResNet-50 \cite{Xiao2018baseline}) & 22.1  & 24.8  & 23.7  & 24.1  & 24.8  & 21.8  & 22.6  & 30.5  & 23.1  & 29.6  & 21.7  & 26.1  & 24.6 \\
    \midrule
    Baseline (ResNet-152 \cite{Xiao2018baseline}) & 19.3  & 21.6  & 22.1  & 23.2  & 22.4  & 20.3  & 18.7  & 28.4  & 21.7  & 28.6  & 20.8  & 23.6  & 22.6 \\
    \midrule
    HRNet-W48 \cite{sun2019deep} & 17.2  & 18.6  & 21.2  & 19.2  & 18.9  & 19.3  & 18.7  & 25.1  & 20.7  & 24.4  & 18.6  & 19.4  & 20.1 \\
    \midrule
    VitPose\cite{xu2022vitpose} & 14.1  & 15.1  & 18.6  & 15.7  & 15.9  & 16.8  & 15.5  & 22.7  & 18.1  & 22.3  & 14.9  & 16.1  & 17.2 \\
    \bottomrule
    \end{tabular}%
  \label{tab:tablempjpe}}%
\end{table*}%

\begin{table}[h]
  \centering
  \caption{This table displays the average PCKh@0.5 values for four different pose estimation models: ResNet50 baseline, ResNet152 baseline, HRNet, and ViTPose. Each model's PCKh@0.5 values are calculated across all key points categories and activities in the dataset.}
    \begin{tabular}{|l|r|}
    \toprule
    \textbf{Methods} & \multicolumn{1}{l|}{PCKh@0.5} \\
    \midrule
    \textit{Resolution: 640x480} &  \\
    \midrule
    Baseline (ResNet-50) & 72.1 \\
    \midrule
    Baseline (ResNet-152) & 74.6 \\
    \midrule
    HRNet-W48 & 78.2 \\
    \midrule
    VitPose & 83.1 \\
    \bottomrule
    \end{tabular}%
  \label{tab:tablepck}%
\end{table}%

\subsection{Performance Evaluation of Pose Estimation Models}

The evaluation of various pose estimation models was conducted on our proposed dataset, employing two standard metrics: Mean Per Joint Position Error (MPJPE) and Percentage of Correct Keypoints (PCKh). Table \ref{tab:tablempjpe} summarizes the MPJPE errors for each activity category, along with the average MPJPE across all activities. Notably, the MPJPE of ViTPose \cite{xu2022vitpose} exhibited a considerable reduction compared to the baseline ResNet-50 model \cite{Xiao2018baseline}, indicating superior accuracy in pose estimation. Further, PCKh@0.5 metric was used for evaluation. Table \ref{tab:tablepck} shows the  PCKh value for different models. There is a noticeable enhancement in PCKh@0.5 metrics with the ViTPose \cite{xu2022vitpose} model, indicating improved keypoint localization capabilities with attention networks.

However, it was observed that poses characterized by self-occlusion, such as sitting and talking, yielded larger MPJPE values compared to poses with minimal occlusion. This underscores the inherent complexity of our dataset, particularly with respect to self-occlusion and occlusion by external objects. Consequently, there exists substantial room for the development of models specifically tailored to address such challenging poses present in our dataset.

\subsection{Visual Analysis of Pose Estimation Results}

Figure \ref{fig:DL_results} presents the visual results obtained from different pose estimation models applied to our proposed dataset. While the models were able to extract relevant features and formulate the pose structure, their performance fell short of expectations compared to their performance on RGB images. Additionally, Figure 2 showcases the failed cases of ViTPose \cite{xu2022vitpose} on images characterized by high occlusion levels.

The challenges encountered in pose estimation on Long Wave Infrared (LWIR) images can be attributed to the inherent differences in texture and contrast compared to RGB images. LWIR images depict thermal intensity rather than visual appearance, rendering occluded poses particularly challenging to identify. Consequently, accurate pose estimation in LWIR images is hindered by the distinct characteristics of thermal imagery.

The evaluation results shed light on the performance and limitations of various pose estimation models when applied to LWIR images in our dataset. These insights guide future research endeavours to improve pose estimation accuracy in challenging conditions characterized by self-occlusion and occlusion in LWIR imagery.
\section{Conclusion}
\label{sec:conclusion}
This paper introduced a first-of-its-kind, fully annotated thermal image dataset for 2D human pose estimation. Featuring over 2400 high-resolution images, it fosters research in this under-explored domain. We evaluated state-of-the-art 2D pose models, revealing their limitations on thermal data but demonstrating significant performance boosts after fine-tuning on our dataset. This establishes a strong baseline for future research. The extensive dataset opens exciting avenues for developing domain-specific models, exploring data fusion, and venturing into tasks beyond pose estimation. By unlocking the potential of thermal data, this work sets the stage for advancements in diverse applications demanding robust pose estimation under challenging conditions.

\bibliographystyle{IEEEbib}
\bibliography{strings,refs}

\end{document}